\documentclass[letterpaper]{article} 
\usepackage[submission]{aaai2026}  
\usepackage{times}  
\usepackage{helvet}  
\usepackage{courier}  
\usepackage[hyphens]{url}  
\usepackage{graphicx} 
\usepackage{natbib}  
\usepackage{caption} 
\usepackage{adjustbox}
\usepackage{graphicx}
\usepackage{subcaption}
\usepackage{booktabs}
\usepackage{algorithm}
\usepackage{algorithmic}
\usepackage{booktabs}
\usepackage{multirow}
\usepackage{xspace}
\usepackage{xcolor}
\usepackage{amsmath}
\usepackage{amssymb}
\usepackage{placeins}
\makeatletter
\DeclareRobustCommand\onedot{\futurelet\@let@token\@onedot}
\def\@onedot{\ifx\@let@token.\else.\null\fi\xspace}

\def\ie{\emph{i.e}\onedot}

\let\showauthors@on=T 
\makeatother
\usepackage{newfloat}
\usepackage{listings}
\DeclareCaptionStyle{ruled}{labelfont=normalfont,labelsep=colon,strut=off} 
\floatstyle{ruled}
\newfloat{listing}{tb}{lst}{}
\floatname{listing}{Listing}
\title{Vision without Images: End-to-End Computer Vision from Single Compressive Measurements}
\author {
    Fengpu Pan\textsuperscript{\rm 1},
    Heting Gao\textsuperscript{\rm 2},
    Jiangtao Wen\textsuperscript{\rm 3},
    Yuxing Han\textsuperscript{\rm 1},
}
\affiliations {
    \textsuperscript{\rm 1}Tsinghua University\\
    \textsuperscript{\rm 2}Southern University of Science and Technology\\
    \textsuperscript{\rm 3}New York University\\
    pfp23@mails.tsinghua.edu.cn, 12210816@mail.sustech.edu.cn, jw9263@nyu.edu, yuxinghan@sz.tsinghua.edu.cn
}

\begin{document}

\maketitle

\begin{abstract}
Snapshot Compressed Imaging (SCI) offers high-speed, low-bandwidth, and energy-efficient image acquisition, but remains challenged by low-light and low signal-to-noise ratio (SNR) conditions. Moreover, practical hardware constraints in high-resolution sensors limit the use of large frame-sized masks, necessitating smaller, hardware-friendly designs. In this work, we present a novel SCI-based computer vision framework using pseudo-random binary masks of only 8×8 in size for physically feasible implementations. At its core is CompDAE, a Compressive Denoising Autoencoder built on the STFormer architecture, designed to perform downstream tasks—such as edge detection and depth estimation—directly from noisy compressive raw pixel measurements without image reconstruction. CompDAE incorporates a rate-constrained training strategy inspired by BackSlash to promote compact, compressible models. A shared encoder paired with lightweight task-specific decoders enables a unified multi-task platform. Extensive experiments across multiple datasets demonstrate that CompDAE achieves state-of-the-art performance with significantly lower complexity, especially under ultra-low-light conditions where traditional CMOS and SCI pipelines fail.
\end{abstract}


\section{Introduction}
\label{sec:intro}

Traditional computer vision tasks including video classification~\cite{Karpathy14}, object detection and tracking~\cite{Wang13,Girshick14}, and monocular depth estimation~\cite{Birkl23midas,Piccinelli24,Bochkovskii2024} are typically performed in the RGB domain using spatial and temporal samples captured by CMOS active pixel sensors (APS). These systems rely heavily on high-throughput image signal processing (ISP) pipelines for denoising~\cite{guo2019agem,li2021denoiser}, demosaicing, and enhancement, resulting in high power consumption, significant latency, and bandwidth bottlenecks. The growing resolution of sensors exacerbates computational burdens in conventional vision pipelines, similar to ``looking for needles in a haystack''. This motivates moving toward new imaging modalities such as compressive sensing \cite{Candes05,Cands2006,Baraniuk2008} based Snapshot Compressive Imaging (SCI~\cite{Hitomi2011,Reddy2011,Llull13,Gao2014,Yuan2016}), where temporal encoding via spatially varying masks allows multiple high-speed frames to be collapsed into a single 2D measurement, enabling dramatic reductions in bandwidth and energy consumption. SCI, however, still faces a fundamental hardware feasibility barrier to adoption. SCI systems typically require large, frame-sized spatial modulation masks to satisfy the Restricted Isometry Property (RIP). One approach to implementing masks is to load them from external memory and project the patterns onto the pixel array prior to each exposure~\cite{Gulve22}. However, when the masks are large, this method introduces substantial latency, power consumption, and system complexity. A more efficient alternative is to directly control pixel states via dedicated external signal lines~\cite{Gulve22}. Yet, as control lines are typically 0.05–1${mu}$m wide and separated by similar spacing, due to physical constraints, each pixel can accommodate at most around 10 such lines. This imposes a stringent upper limit on mask size, making large dynamic masks infeasible for practical sensor integration.

Meanwhile, low-light and low signal-to-noise ratio (SNR) conditions are particularly challenging for vision systems based on both ASPs and SCI, yet critically important in scenarios such as autonomous driving and robotics. These conditions give rise to complex noise patterns, commonly modeled as Poisson-Gaussian noise~\cite{Foi08}. In dynamic scenes, where signal intensity fluctuates over time, the Poisson noise becomes non-stationary and temporally correlated \cite{ottewill1997correlation,sun2024}, deviating from the conventional memoryless assumption. Existing denoising neural networks~\cite{Khademi21} rarely account for such temporal dependencies, and robust computer vision under extreme low-light remains an open challenge~\cite{dong2010fast}. 

To overcome these issues, we propose an end-to-end SCI-based vision system that operates directly on raw measurements using pseudo-random binary spatial modulation masks that are only 8×8 in size, much smaller than conventionally needed for SCI and well-suited to existing CMOS implementations, enabling low-latency, low-power vision at the sensor level. Instead of reconstructing full video frames then applying computer vision algorithms using reconstructed full frames~\cite{Kwan19}, we design a unified model that performs vision tasks directly in the compressive measurement domain. While some prior works have explored reconstruction-free inference~\cite{Bethi21,Hu21,Zhang22,Lu20cav,okawara20}, most directly apply standard vision models to mosaic-like SCI data, leading to suboptimal results. For instance, in~\cite{Lu20cav}, high-speed object motion in compressed highway footage led to recognition failures due to motion artifacts. Our proposed framework, CompDAE, on the other hand, leverages a Transformer-based encoder-decoder architecture to learn spatiotemporal features directly from noisy compressive measurements. The same encoder can be paired with different lightweight decoders for edge detection, depth estimation, or semantic segmentation—achieving both hardware and computational efficiency. In training the Transformer, we incorporate a rate-constrained training strategy inspired by the recent BackSlash algorithm~\cite{wu2025backslashrateconstrainedoptimized}, leading to edge-deployment-friendly models that are model compression and pruning friendly, enabling low-complexity, high-efficiency vision tasks such as edge detection and depth estimation that are robust to a wide range of lighting and SNR conditions. Extensive evaluations across multiple datasets and tasks demonstrate that CompDAE consistently outperforms conventional APS and SCI-based pipelines, particularly under ultra-low-light conditions where existing methods fail. 

\section{Related work}
\label{sec:relate}

\noindent\textbf{Video Snapshot Compressed Imaging (SCI)} is a computational imaging technique that captures full frame rate video in a single 2D measurement by modulating each frame with a distinct mask and summing them. Reconstruction algorithms are then used to recover the original frames based on compressive sensing. Various CMOS-compatible designs have been proposed to enable on-chip SCI. For example, Dadkhah et al.\cite{Dadkhah14} introduced a block-based CS method using a three-transistor APS and 8×8 pseudo-random exposure, achieving a compact, low-power implementation but with limited flexibility and high reconstruction complexity under low-photon conditions. Yoshida et al.\cite{Yoshida20} developed quasi pixel-wise exposure control within 8×8 blocks, though its fixed structure hampers noise cancellation in photon-limited settings. Zhang et al.\cite{Compact} implemented pixel-wise coded exposure entirely in CMOS with in-pixel memory, enabling ultra-low power operation but limiting adaptability due to a fixed single-on pattern. More recently, Liu et al.\cite{Liu:24} proposed an adaptive optical architecture with dynamic mask selection, improving performance across lighting conditions at the cost of increased system size and photon loss. In contrast, our approach adopts a fixed-mask optical design and introduces a compressive denoising autoencoder built on spatiotemporal Transformers, enabling robust, high-quality video understanding directly from noisy compressive measurements, even under severe photon-limited conditions.

Recent advances in SCI have extended encoding strategies beyond simple binary masks. Multi-bit and multi-state schemes—such as Gray codes and progressive masks—offer greater dynamic range to support both low-light and normal-light conditions. Wang et al.~\cite{Wang2023Structural} showed that deep learning can adapt encoding parameters in real time to varying illumination using high-refresh-rate digital mirror devices (DMDs). In parallel, more accurate noise modeling has emerged, with additive Gaussian~\cite{Inagaki18} and Poisson~\cite{Liu:24} models better capturing low-light sensor behavior. End-to-end optimization of learnable masks has also been explored~\cite{CoCoCs, Yoshida18, iliadis2020deepbinarymask}. In contrast, our method uses fixed 8×8 repeated binary masks on a DMD, simplifying hardware implementation while providing reliable, consistent modulation.

In video analysis, reconstruction-based methods first recover full video frames from compressive measurements before applying conventional frame-by-frame algorithms. For instance, Hu et al.~\cite{Hu2021} perform object detection on reconstructed video, Lu et al.~\cite{Lu20cav} process restored footage for sequential tasks, and Kwan et al.~\cite{Kwan19} use reconstructed images for target tracking. In contrast, reconstruction-free methods operate directly on compressive measurements: Liang et al.~\cite{liang2022} and Zhang~\cite{Zhang22} extract semantic features without full reconstruction, while Bethi et al.~\cite{Bethi21} and Lu et al.~\cite{Lu20cav} conduct recognition tasks directly on compressed inputs. Our approach follows this latter direction, but goes further by extracting spatiotemporal semantic representations end-to-end, eliminating the need for explicit reconstruction entirely.

Reconstructing video from SCI measurements is an ill-posed inverse problem. Traditional model-based methods rely on iterative optimization~\cite{Liao14,Boyd11} with handcrafted priors such as total variation~\cite{Yuan16} or non-local low-rank structures~\cite{Liu18}, but often suffer from poor quality and high computational cost. These limitations have led to deep learning-based approaches: GAP-net~\cite{Meng20}, DUN-3DUnet~\cite{Wu2021}, and Two-stage~\cite{Zheng2022TwoStage} leverage deep unfolding for better speed and accuracy. STFormer~\cite{Wang22} introduces space-time factorization and local self-attention to model spatiotemporal correlations efficiently. While Transformers~\cite{Vaswani17} capture long-range dependencies, their high complexity poses challenges for large-scale color video. EfficientSCI~\cite{Wang23} addresses this by combining CNNs and Transformers with hierarchical residuals, achieving strong performance on UHD color video reconstruction.

\noindent\textbf{Denoising Autoencoders (DAEs)} learn robust representations by reconstructing clean inputs from corrupted versions~\cite{Vincent08}. Given a noisy input $x_{noisy}$, the encoder maps it to a latent code $z=f(x_{noisy})$ which the decoder then uses to recover the original signal $x_{clean}$, DAEs are widely used in vision tasks for extracting noise-invariant features~\cite{Pathak16,Chen20,Dosovitskiy20,He22}. Recent works apply Transformers~\cite{Vaswani17} to unify vision and language modeling. iGPT~\cite{Chen20} treats pixels as tokens, while ViT~\cite{Dosovitskiy20} uses patches, enabling masked prediction. MAE~\cite{He22} builds on DAE principles with a masked autoencoding framework that reconstructs missing image patches from sparse visible inputs using an asymmetric encoder-decoder design, capturing high-level semantics efficiently.

\noindent\textbf{Rate-Constrained Training}, inspired by BackSlash~\cite{wu2025backslashrateconstrainedoptimized}, introduces a regularization term based on Exp-Golomb coding to balance model accuracy and parameter compressibility. We adopt this strategy during training to guide the network toward compact, efficient representations, supporting both reconstruction and downstream vision tasks in our compressive sensing framework.

\section{Single Measurement SCI-based Computer Vision in Low-Lighting Conditions}
\label{sec:method}
\subsection{Imaging Pipeline and Noise}
\label{formula}
For gray-scale video SCI, we use coded aperture compressive temporal imaging (CACTI)~\cite{Llull13} as an example. The frames of a video sequence are modulated temporally and spatially and then compressed to a single measurement. The number of coded frames for a single measurement is determined by the mask or different patterns on the DMD within the integration (exposure) time. Specifically, consider that $T$ video frames are modulated by $T$ different modulation patterns. Let $\{\mathbf{X}_t\}^T_{t=1} \in \mathbb{R}^{n_x \times n_y}$ denote a $T$-frame high-speed scene to be captured in a single exposure time, where $n_x$, $n_y$ represent the spatial resolution of each frame and $T$ is the compression ratio (Cr) of the video SCI system. Then, the modulation process can be modeled as multiplying $\{\mathbf{X}_t\}^T_{t=1}$ by pre-defined masks $\{\mathbf{M}_t\}^T_{t=1} \in \mathbb{R}^{n_x \times n_y}$, 
\begin{equation}
\mathbf{Y}_t = \mathbf{X}_t \odot \mathbf{M}_t,
\end{equation}
where $\{\mathbf{Y}_t\}^T_{t=1} \in \mathbb{R}^{n_x \times n_y}$ and $\odot$ denote the modulated frames and element-wise Hadamard product, respectively. Modulated frames are then summed to a single measurement ${\mathbf{Y}} \in \mathbb{R}^{n_x \times n_y}$. 
Thus, the forward model of the video SCI system can be formulated as, 
\begin{equation}
    \mathbf{Y} = \sum^T_{t=1} {\mathbf{X}_t \odot \mathbf{M}_t} + \mathbf{N},
\end{equation}
where $\mathbf{N} \in \mathbb{R}^{n_x \times n_y}$ denotes the measurement noise. 
In the context of photon-limited applications (such as low-light imaging), the Poisson-Gaussian model is generally used for raw-data digital imaging sensor data~\cite{Foi08}. In general, the noise term is composed of two mutually independent parts, a Poisson signal-dependent component $\eta_p$ and a Gaussian signal-independent component $\eta_g$. 
\begin{equation} \label{eq:2}
    \mathbf{Y_{noisy}} = \sum^T_{t=1} {\left(\mathbf{X}_t \odot \mathbf{M}_t + \eta_{p,t}\right)} + \eta_{g}.
\end{equation}
The distributions of these two components are characterized as follows, $\alpha (\mathbf{X}_t \odot \mathbf{M}_t + \eta_{p,t}) \sim \mathcal{P}(\alpha (\mathbf{X}_t \odot \mathbf{M}_t))$, $\eta_g \sim \mathcal{N}(0, \sigma)$, where $\alpha>0$ and $\sigma>0$ are real scalar parameters and $\mathcal{P}$ and $\mathcal{N}$ Denote the Poisson and normal distributions.

\subsection{CompDAE: Visual Intelligence from Raw Temporal Measurements}
At the core of our proposed SCI-based compressive computer vision system is CompDAE, a DAE~\cite{Vincent08} that reconstructs the original clean spacetime data from its raw temporal compressive measurement captured by SCI sensors. Like all DAEs, our approach consists of an encoder that maps noisy raw measurements into a latent representation, and a decoder that reconstructs the original spacetime signal from this latent representation. Similar to the Masked Autoencoder (MAE)~\cite{He22}, we adopt an asymmetric design where the encoder operates both the measurement and modulation masks of the SCI system, while a lightweight decoder (with $N$\textless$M$/2) reconstructs the full signal from the latent representation. 

\begin{figure}[!htbp]
    \centering
    \includegraphics[width=0.8\linewidth]{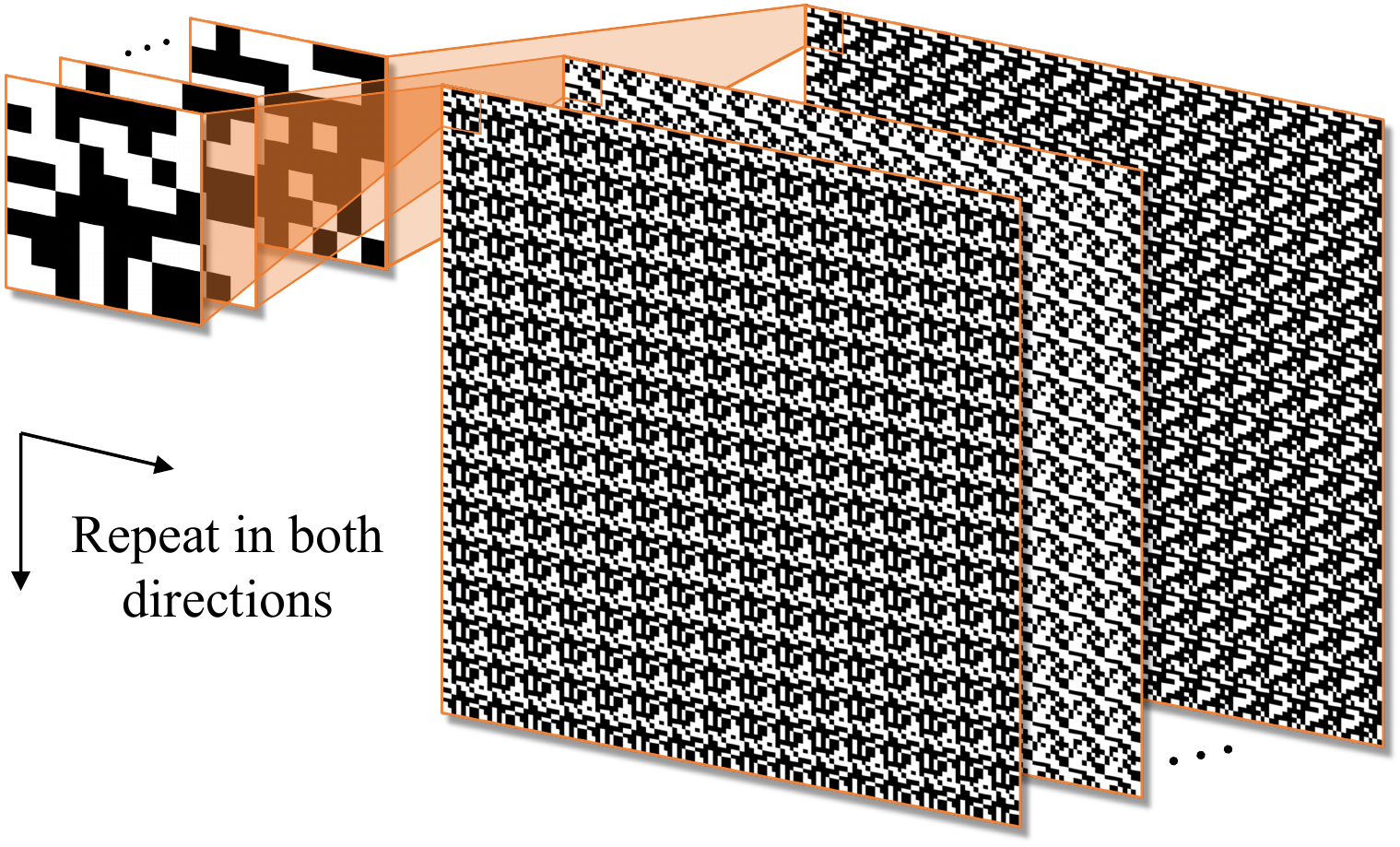}
    \caption{Construction of binary mask pattern.}
    \label{fig:binary-mask}
\end{figure}

\noindent\textbf{Hardware-friendly mask design}
Due to physical constraints, the masks or DMD responses to light are always nonnegative and bounded~\cite{Llull13,Jiang15}. 
Moreover, practical SCI systems often employ binary-valued masks. Theoretical analysis shows that i.i.d. binary (Bernoulli) masks are feasible for signal recovery~\cite{Zhao23}, while the probability of non-zero entries $\rho$ is related to reconstruction distortion and a hyperparameter in mask design. As illustrated in Figure~\ref{fig:binary-mask}, binary masks for the entire frame in CompDAE are generated by spatially replicating smaller binary sub-masks $\{\mathbf{A}_t\}^T_{t=1} \in \{0,1\}^{m_x \times m_y}$ in both directions, i.e. $\mathbf{M}_t = \mathbf{J}_{n_x/m_x} \otimes \mathbf{A}_t$, where $\mathbf{J}_{n_x/m_x}$ is an all-ones matrix of size $n_x/m_x$ and $\otimes$ represents the Kronecker product.

\noindent\textbf{CompDAE encoder and decoder}
Following STFormer, we adopt an estimation module to pre-process the measurement and masks, as in~\cite{Cheng21,Cheng20}. Our encoder consists of a token generation (TG) block using 3D CNNs, followed by stacked STFormer blocks that model spatial-temporal dependencies.
Unlike STFormer, the estimation module comprises a Noise Level Mapping (NLM) for unified noise-adaptive training, and the decoder is used only during pre-training~\cite{Cheng21} for reconstruction. To support diverse downstream tasks efficiently, we design the decoder with a few STFormer blocks followed by a convolutional layer. We adopt a partial fine-tuning strategy~\cite{He22}, updating only the decoder while keeping the encoder fixed. This asymmetric design enables task-specific decoders to be lightweight and fast to train, while still benefiting from shared, robust spatiotemporal representations.

\noindent\textbf{Unified Noise-Adaptive Training Strategy}
To support generalization and real-world deployment, we propose a unified training strategy that enables a single CompDAE model to handle diverse noise levels. The model takes a multi-channel input tensor comprising a signal estimate, a Gaussian noise map, and a Poisson noise proxy via Average Photon Count (APC). This design allows the network to learn noise-aware features and adapt dynamically to varying lighting and SNR conditions.

\noindent\textbf{Training-Time Compression (BackSlash)} 
We apply rate-constrained training (RCT), inspired by BackSlash~\cite{wu2025backslashrateconstrainedoptimized}, to balance performance and model size. Using exponential-Golomb (EG) codes~\cite{wen1999structured}, we regularize parameter sparsity by minimizing $J(\theta) = L_{\text{task}}(\theta) + \lambda \, R(\theta)$, where \(L_{\text{task}}(\theta)\) denotes the task-specific loss and the rate term is defined as $R(\theta) = \frac{1}{N} \sum_{i=1}^{N} \left( |\theta_i| + \epsilon \right)^{\nu} $, with $N$ being the total number of parameters, \(\epsilon\) a small constant, and \(\nu\) the GG shape parameter. This explicit rate regularization encourages the compressibility of GG-distributed parameters. 
CompDAE learns directly from noisy, raw SCI measurements under varying SNRs. It generates tokenized video representations during pre-training, then fine-tunes lightweight decoders for pixel-level reconstruction, classification, or regression, supporting modular vision tasks under low-light conditions.

\section{Experiment Results}
\label{sec:experiment}

We evaluate CompDAE through self-supervised pre-training on the DAVIS2017 dataset~\cite{Pont17}, using 90 videos (6242 frames at 480×894 resolution). We then apply transfer learning for two vision tasks: (i) edge detection and (ii) monocular depth estimation.
For edge detection, high-quality ground truth maps are generated using DexiNed~\cite{Poma20DexiNed}, and the fine-tuned model is tested on six gray-scale benchmark datasets~\cite{Yuan20PnP}. For depth estimation, we use RGB-guided depth completion on KITTI~\cite{Uhrig17} to train and evaluate the model on simulated raw SCI measurements.
We further test segmentation on Cityscapes~\cite{Cordts2016Cityscapes}, where dense annotations exist only for sparse frames. Leveraging CompDAE’s temporal modeling, we fine-tune the segmentation decoder using limited training annotations, demonstrating strong generalization across time.

\subsection{Implementation Details}
\label{default}

We train our model using PyTorch on a single NVIDIA A100 (80GB) GPU. Photon-limited measurements are generated following the process in Sec.~\ref{formula}, with Gaussian noise ($\sigma = 0.01$) added by default to simulate sensor noise. Data augmentation includes random cropping, flipping, and APC variation. Training uses the Adam optimizer~\cite{Kinga15adam} with an initial learning rate of 0.0001. For compressive sampling, we use 8-frame sequences at 128×128 resolution (Cr = 8), sampled from the original sequence using a stride of 1 or 2 to preserve temporal redundancy.
 
Training consists of two stages: self-supervised pre-training on noisy measurements (80 epochs) and task-specific fine-tuning. For edge detection, we fine-tune on 128×128 inputs for 40 epochs. For depth estimation, the KITTI data is resized to 256×256 and fine-tuned for 50 epochs. For semantic segmentation, we fix the encoder and train a PPM head~\cite{Zhao2017pspnet} on color Bayer measurements from Cityscapes (256×512) for 40 epochs.

We evaluate CompDAE using task-specific metrics. For edge detection, we report Optimal Dataset Scale (ODS) and Optimal Image Scale (OIS) F1-scores. Depth estimation performance is measured by AbsRel, RMSE, log10 error, and accuracy thresholds ($\delta_1$, $\delta_2$, $\delta_3$). For segmentation, we use mean Intersection over Union (mIoU) and mean Pixel Accuracy (mPA).

\subsection{Comparison with conventional systems}

\begin{figure}[!h]
    \centering
    \includegraphics[width=1.0\linewidth]{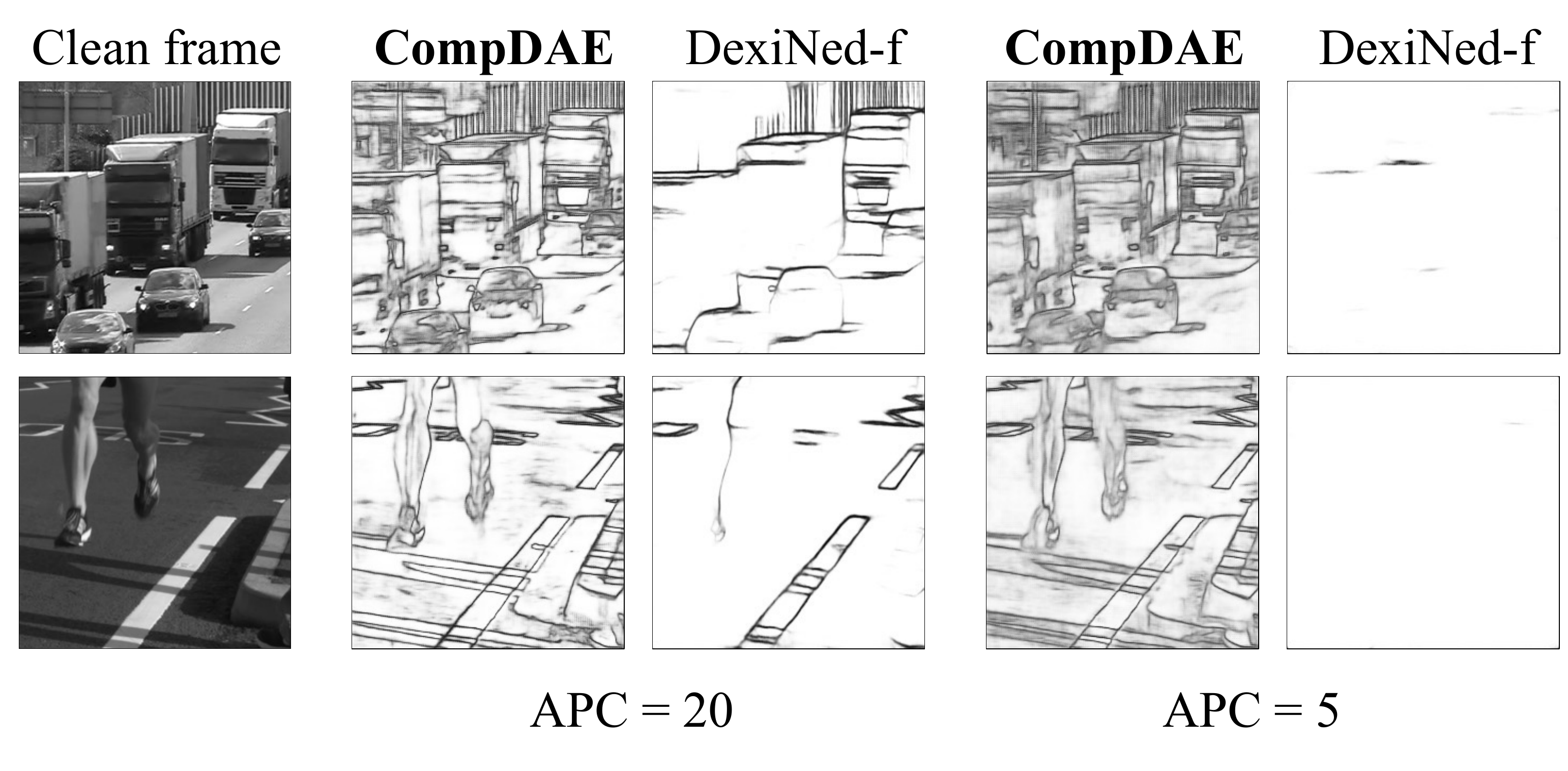}
    \caption{Comparison of edge detection results under different ultra-low-lighting conditions.}
    \label{fig:ultra-low-lighting}
\end{figure}

We highlight the failure of conventional edge detection algorithms under ultra-low-light conditions. As shown in Figure~\ref{fig:ultra-low-lighting}, DexiNed struggles at low APC, producing noisy, indistinct edges, while CompDAE maintains clear structure. We further compare CompDAE with classical edge detectors (Canny~\cite{Canny1986}, RCF~\cite{Liu17RCF}, BDCN~\cite{He19BDCN}) and depth estimation methods (MiDaS~\cite{Birkl23midas}, Unidepth~\cite{Piccinelli24}, Metric3D~\cite{hu2024metric3dv2}, LightedDepth~\cite{zhu2023lighteddepth}). Quantitative comparisons are given in Table~\ref{tab:edge_detection}, which shows that CompDAE outperforms conventional APS-based methods (Canny, RCF, BDCN, DexiNed) in ODS and OIS scores at APC = 20. Unlike frame-by-frame ISP-based approaches, CompDAE processes a single low-SNR measurement, yet still achieves superior edge detection under severe low-light conditions.

CompDAE effectively maintains high performance under low-light conditions.

\renewcommand{\arraystretch}{0.8}
\begin{table}[!htbp]
  \centering
  \scriptsize
  \begin{tabular}{@{}lccc}
    \toprule
    Method & ODS & OIS & Params (M) \\
    \midrule
    Canny& 0.450 & 0.455 & - \\
    BDCN& 0.389& 0.441& 16.30\\
    RCF& 0.427& 0.454&  14.80\\
    DexiNed& 0.474 & 0.479& 35.22 \\
    \midrule
    CompDAE &  \textbf{0.690}&  \textbf{0.704}& 23.82 \\
    \bottomrule
  \end{tabular}
  \caption{Evaluation on gray-scale video benchmark datasets with our fine-tuned CompDAE.  
  }
  \label{tab:edge_detection}
\end{table}

\renewcommand{\arraystretch}{0.8}
\begin{table}[!htbp]
  \centering
  \scriptsize
  \begin{tabular}{@{}lcccccc@{}}
    \toprule
    Method & AbsRel & RMSE & $\log${10} & $\delta_1$ & $\delta_2$ & $\delta_3$ \\
    \midrule
    Unidepth &  0.331&  7.483&  0.353&  0.561&  0.829&  0.919\\
    MiDaS &  0.431&  8.116&  1.658&  0.352&  0.642& 0.812 \\
    Metric3D &  0.329&  8.600&  0.351&  0.582&  0.843&  0.922\\
    LightedDepth &  0.361&  12.016&  0.743&  0.521&  0.635&  0.731\\
    \midrule
    CompDAE$_{N=0}$& 0.309& 7.203& 0.368& 0.670& 0.812& 0.894\\
    CompDAE$_{N=1}$& \textbf{0.181}& \textbf{5.177}& \textbf{0.262}& \textbf{0.770}& \textbf{0.893}& \textbf{0.951}\\
    \bottomrule
  \end{tabular}
  \caption{Evaluation on KITTI with fine-tuned CompDAE.}
  \label{tab:depth_estimation1}
\end{table}

Table~\ref{tab:depth_estimation1} further shows that CompDAE with a decoder depth of $N$ = 1, outperforms Unidepth and Metric3D in depth estimation under the same low-light setting, demonstrating its strong generalization across vision tasks. Notably, while Unidepth performs well at APC = 20, it requires 353.83M parameters, 15 times that of our model. Despite this, CompDAE ($N$ = 0) achieves comparable accuracy with far lower computational cost. Moreover, CompDAE boasts an inference time of 2.60 ms per frame, vastly outperforming Unidepth’s 63.90 ms, making CompDAE far more suitable for real-time applications.

\begin{figure}[t!]
    \centering
    \begin{minipage}{0.325\columnwidth}
        \centering
        \footnotesize Clean Frame \\ \vspace{2pt}
        \includegraphics[width=\linewidth]{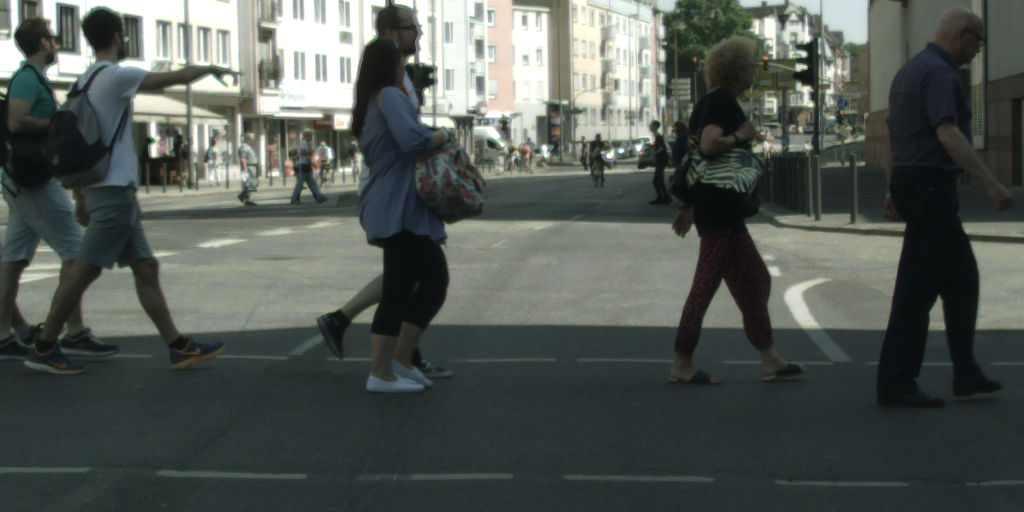}
    \end{minipage}
    \hfill
    \begin{minipage}{0.325\columnwidth}
        \centering
        \footnotesize Noisy (APC = 20) \\ \vspace{2pt}
        \includegraphics[width=\linewidth]{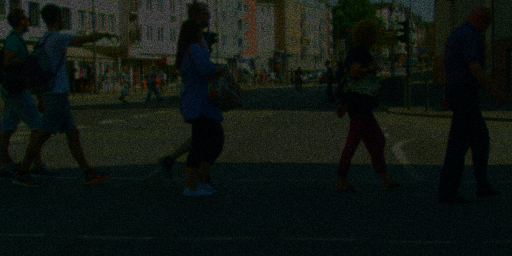}
    \end{minipage}
    \hfill
    \begin{minipage}{0.325\columnwidth}
        \centering
        \footnotesize GT \\ \vspace{2pt}
        \includegraphics[width=\linewidth]{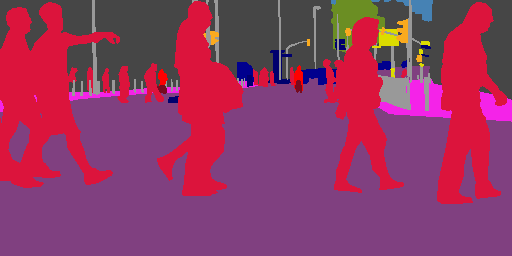}
    \end{minipage}

    \vspace{0.5em}
    \begin{minipage}{0.325\columnwidth}
        \centering
        \footnotesize SegFormer \\ \vspace{2pt}
        \includegraphics[width=\linewidth]{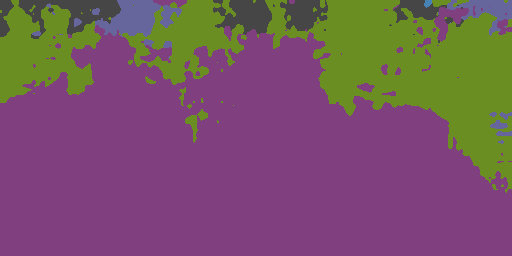}
    \end{minipage}
    \hfill
    \begin{minipage}{0.325\columnwidth}
        \centering
        \footnotesize DIC+SegFormer \\ \vspace{2pt}
        \includegraphics[width=\linewidth]{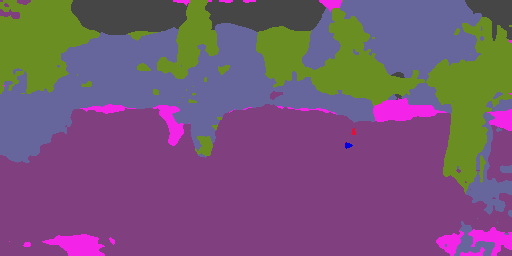}
    \end{minipage}
    \hfill
    \begin{minipage}{0.325\columnwidth}
        \centering
        \footnotesize \textbf{CompDAE} \\ \vspace{2pt}
        \includegraphics[width=\linewidth]{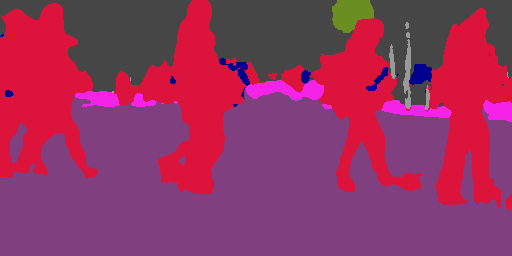}
    \end{minipage}

    \caption{Qualitative comparison for semantic segmentation on Cityscapes.}
    \label{fig:semantic-segmentation}
\end{figure}

To assess the encoder’s ability to capture semantic and temporal features beyond edges, we perform semantic segmentation using raw Bayer measurements. Given limited annotated video data, we extract features from a single annotated frame and fine-tune only the decoder. As shown in Figure~\ref{fig:semantic-segmentation}, our method produces more accurate and plausible segmentation maps than the baseline. At APC = 20, CompDAE achieves 30.43\% mIoU and 82.49\% mPA, significantly outperforming SegFormer (6.65\% mIoU, 58.09\% mPA), demonstrating strong semantic understanding under compressive sensing constraints.

\subsection{Ablation study observations}
We conduct controlled ablation studies by modifying key components of CompDAE, using shorter training schedules for broader experimentation. Unless noted, both NLM and BackSlash are excluded. The results reveal several notable insights.

As mentioned earlier, decoder depth $N$ refers to the number of STFormer blocks in the decoder. Within our system framework, we use partial fine-tuning~\cite{Zhang16,He22} to optimize efficiency. 
Our asymmetric setup (\ie,$N$\textless$M$/2 with $M$ = 4) keeps the decoder lightweight and adaptable. As shown in Table~\ref{tab:depth}, increasing $N$ initially improves ODS and OIS scores, peaking at depth 2. Further increasing depth to 3 slightly degrades performance, suggesting diminishing returns relative to added complexity.

Table~\ref{tab:compression ratio} shows the impact of compression ratio (Cr).
A higher Cr reduces data bandwidth but increases compression. We find that Cr = 8 offers the best trade-off, preserving scene dynamics while minimizing information loss, particularly under low-light, low-SNR conditions where traditional methods struggle. Consistent Cr values during training and testing yield optimal performance, while mismatches significantly degrade results. These findings highlight the importance of aligning Cr settings to maintain accuracy and preserve the system’s low-complexity, high-speed benefits.

Table~\ref{tab:Mask pattern} examines the impact of mask density, defined by $\rho$, the probability of non-zero entries in the 8×8 binary masks. While higher $\rho$ allows more light and potentially increases photon counts, it may reduce spatial modulation. Our results show that within the 0.3-0.5 range, $\rho$ has minimal effect on reconstruction quality, even under low-light conditions. This supports findings in~\cite{iliadis2020deepbinarymask} that smoother sensing patterns can maintain performance regardless of $\rho$, highlighting the robustness of our system across different mask densities.
\renewcommand{\arraystretch}{0.8}
\begin{table*}[t!]
    \centering
    \scriptsize
    \begin{subtable}{.3\linewidth}
        \centering
        \begin{tabular}{@{}lcc@{}} 
            \toprule
            Depth& ODS& OIS \\ 
            \midrule
            $N=0$& 0.672& 0.685 \\  
            $N=1$& 0.681& 0.694\\ 
            $N=2$& 0.695& 0.709\\ 
            $N=3$& 0.686& 0.698\\
            \bottomrule
        \end{tabular}
        \caption{Impact of Decoder Depth.}
        \label{tab:depth}
    \end{subtable}
    \hfill
    \begin{subtable}{.4\linewidth}
        \centering
        \begin{tabular}{@{}lcccccc@{}} 
            \toprule
            \multirow{2}{*}{Cr}& \multicolumn{2}{c}{4} & \multicolumn{2}{c}{8}& \multicolumn{2}{c}{16}\\ 
            \cmidrule(lr){2-3} \cmidrule(lr){4-5} \cmidrule(lr){6-7}
            & ODS& OIS & ODS& OIS & ODS&OIS \\ 
            \midrule
            4& 0.669& 0.682& 0.587& 0.593& 0.459& 0.497\\ 
            8& -& -& 0.672& 0.685& 0.513& 0.538\\ 
            16& -& -& -& -& 0.612&0.617\\ 
            \bottomrule
        \end{tabular}
        \caption{Impact of Compression Ratio (Cr).}
        \label{tab:compression ratio}
    \end{subtable}
    \hfill
    \begin{subtable}{.25\linewidth}
        \centering
        \begin{tabular}{@{}lcc@{}} 
            \toprule
            $\rho$& ODS& OIS \\ 
            \midrule
            0.3& 0.673& 0.680\\ 
            0.4& 0.638& 0.652\\ 
            0.5& 0.672& 0.685\\ 
            \bottomrule
        \end{tabular}
        \caption{Impact of Mask Pattern Density.}
        \label{tab:Mask pattern}
    \end{subtable}
    \caption{Ablation studies on key hyperparameters}
    \label{tab:ablation_combined}
\end{table*}

\renewcommand{\arraystretch}{0.8}
\begin{table*}[t]
\centering
\scriptsize
\begin{tabular}{@{}ll|ccc|ccc|ccc@{}}
\toprule
\multirow{2}{*}{Model Type} & \multirow{2}{*}{Strategy} & \multicolumn{3}{c|}{APC = 5} & \multicolumn{3}{c|}{APC = 10} & \multicolumn{3}{c}{APC = 20} \\
\cmidrule(lr){3-11}
  &  & ODS & OIS & EGCR(\%) & ODS & OIS & EGCR(\%) & ODS & OIS & EGCR(\%) \\ \midrule
\multirow{3}{*}{APC-specific} 
   & w/o BackSlash & 0.660 & 0.677 & --    & 0.676 & 0.691 & --    & 0.696 & 0.712 & --    \\
   & w/ BackSlash  & 0.629 & 0.642 & 83.91 & 0.653 & 0.672 & 83.93 & 0.674 & 0.691 & 83.93 \\
   & half BackSlash & 0.646 & 0.661 & 48.66 & 0.666 & 0.684 & 48.31 & 0.690 & 0.704 & 48.70 \\  \midrule
\multirow{3}{*}{Unified}
   & w/o BackSlash & 0.638      & 0.656      & --     & 0.669      & 0.687      & --     & 0.683      & 0.701      & --     \\
   & w/ BackSlash  & 0.606      & 0.621      & 83.63     & 0.634      & 0.651      & 83.63     & 0.659      & .674      & 83.63   \\
   & half BackSlash  & 0.621      & 0.639      & 50.06     & 0.650      & 0.669      & 50.06     & 0.670      & 0.685      & 50.06  \\ 
\bottomrule
\end{tabular}
\caption{Edge Detection Performance comparison for different training strategies under varying APC test conditions.}
\label{tab:combined_comparison}
\end{table*}

\subsection{Training Strategy Comparisons}
We assess the effect of the BackSlash rate-constrained training strategy on performance. Two model types were tested: APC-specific models trained at fixed APC levels (5, 10, 20) and a unified model trained across a wide APC range (1-60) and evaluated at the same levels. Compression efficiency is measured by Effective Golomb Compression Rate (EGCR), where a lower average Exp-Golomb code length indicates higher compactness. Without BackSlash, both model types already perform well, with the unified model showing strong generalization. However, applying BackSlash throughout training slightly degrades edge detection. To address this, we adopt a Half BackSlash strategy—applied only during the early training phase—which preserves accuracy while improving parameter efficiency. As shown in Table~\ref{tab:combined_comparison}, this approach balances compactness and task performance under varied lighting conditions.

\subsection{Comparison with Other SCI-based Methods}

To highlight the advantages of CompDAE, we compare it against representative methods in two aspects: (1) its superior performance as an end-to-end alternative to traditional two-stage SCI approaches, and (2) its robustness as a multi-task, multi-condition platform. All experiments use our unified model trained with the Half BackSlash strategy for consistent evaluation.

\begin{figure*}[t!]
    \centering
    \setlength{\fboxrule}{0.2pt}
    \setlength{\fboxsep}{0pt}

    \begin{minipage}[c]{0.135\linewidth} \centering \footnotesize Clean Frame \end{minipage}\hfill
    \begin{minipage}[c]{0.135\linewidth} \centering \footnotesize GT \end{minipage}\hfill
    \begin{minipage}[c]{0.135\linewidth} \centering \footnotesize \textbf{CompDAE} \end{minipage}\hfill
    \begin{minipage}[c]{0.135\linewidth} \centering \footnotesize DIC+BDCN \end{minipage}\hfill
    \begin{minipage}[c]{0.135\linewidth} \centering \footnotesize DIC+DexiNed \end{minipage}\hfill
    \begin{minipage}[c]{0.135\linewidth} \centering \footnotesize DIC+RCF \end{minipage}\hfill
    \begin{minipage}[c]{0.135\linewidth} \centering \footnotesize DIC+DDN \end{minipage}
    \\[2pt]

    \fbox{\includegraphics[width=0.126\linewidth]{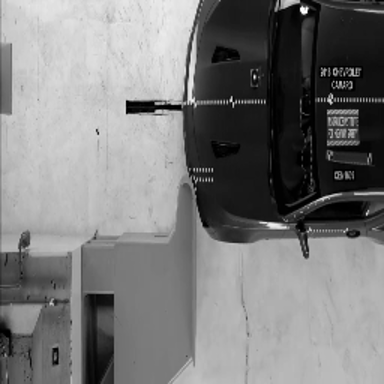}}\hfill
    \fbox{\includegraphics[width=0.126\linewidth]{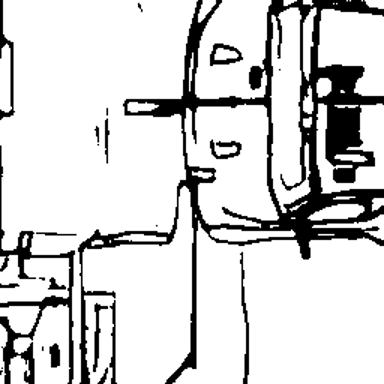}}\hfill
    \fbox{\includegraphics[width=0.126\linewidth]{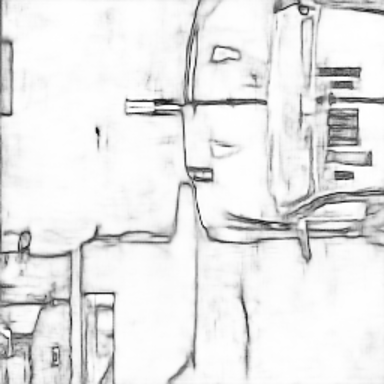}}\hfill
    \fbox{\includegraphics[width=0.126\linewidth]{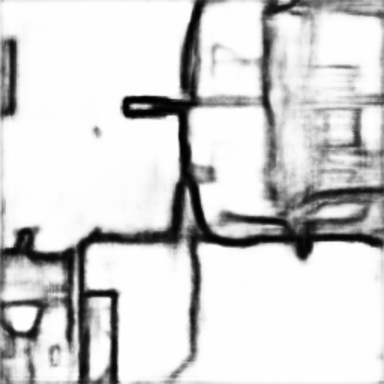}}\hfill
    \fbox{\includegraphics[width=0.126\linewidth]{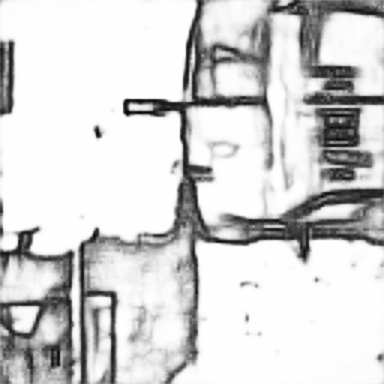}}\hfill
    \fbox{\includegraphics[width=0.126\linewidth]{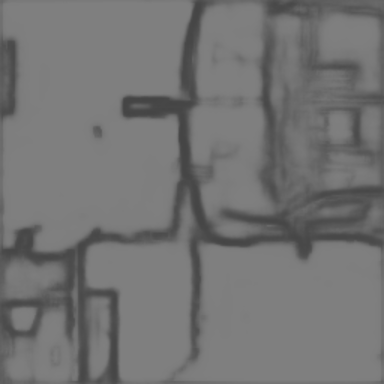}}\hfill
    \fbox{\includegraphics[width=0.126\linewidth]{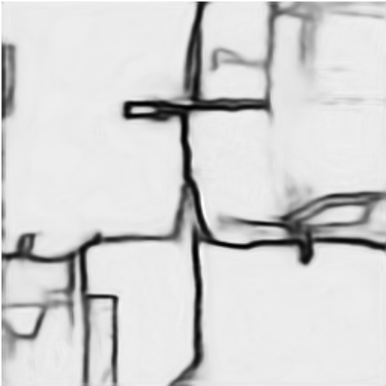}}
    \\[2pt]

    \fbox{\includegraphics[width=0.126\linewidth]{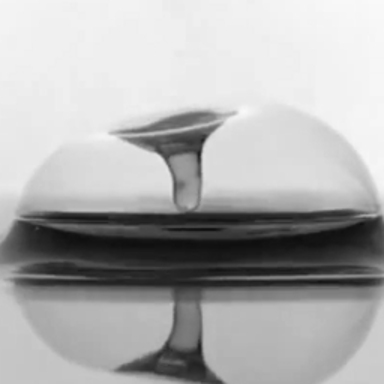}}\hfill
    \fbox{\includegraphics[width=0.126\linewidth]{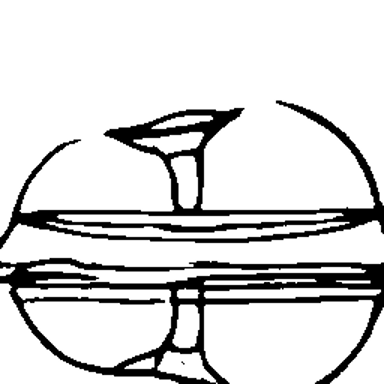}}\hfill
    \fbox{\includegraphics[width=0.126\linewidth]{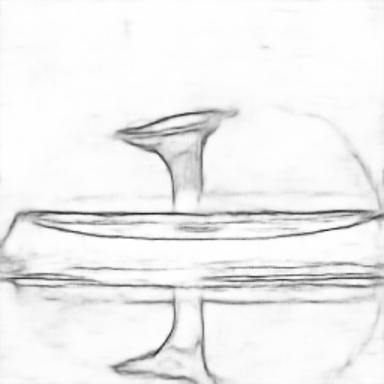}}\hfill
    \fbox{\includegraphics[width=0.126\linewidth]{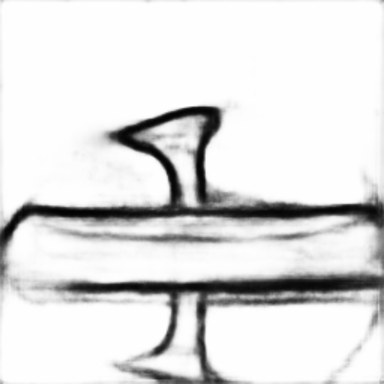}}\hfill
    \fbox{\includegraphics[width=0.126\linewidth]{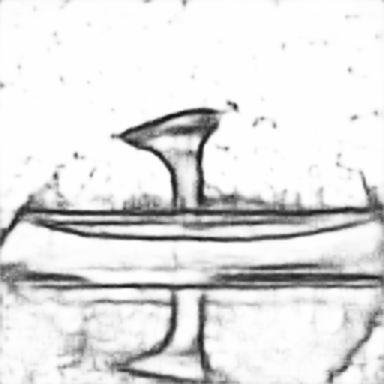}}\hfill
    \fbox{\includegraphics[width=0.126\linewidth]{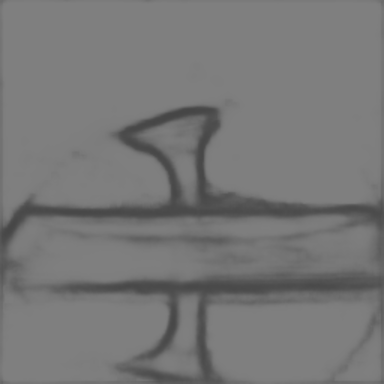}}\hfill
    \fbox{\includegraphics[width=0.126\linewidth]{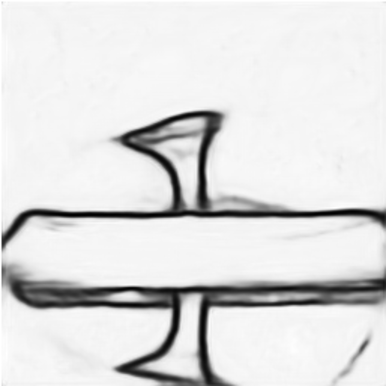}}
    \\[2pt]

    \fbox{\includegraphics[width=0.126\linewidth]{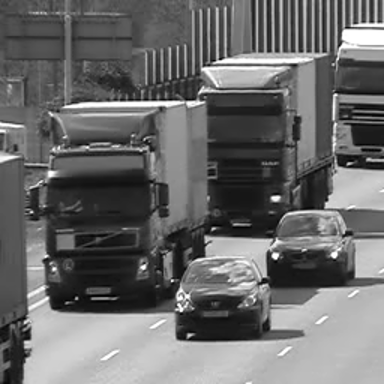}}\hfill
    \fbox{\includegraphics[width=0.126\linewidth]{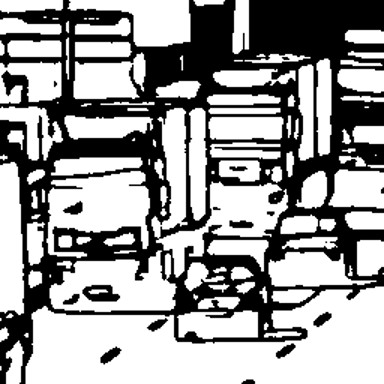}}\hfill
    \fbox{\includegraphics[width=0.126\linewidth]{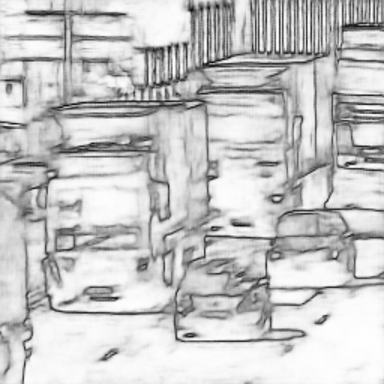}}\hfill
    \fbox{\includegraphics[width=0.126\linewidth]{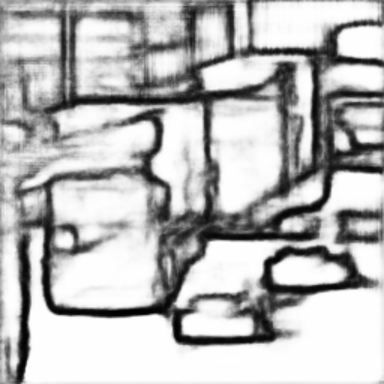}}\hfill
    \fbox{\includegraphics[width=0.126\linewidth]{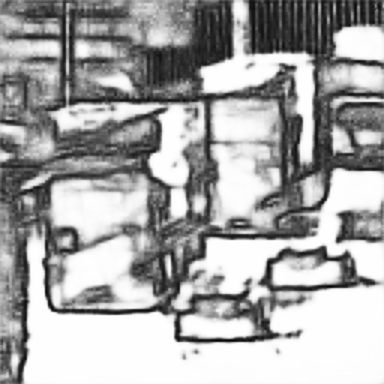}}\hfill
    \fbox{\includegraphics[width=0.126\linewidth]{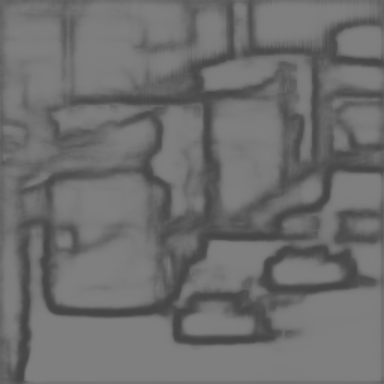}}\hfill
    \fbox{\includegraphics[width=0.126\linewidth]{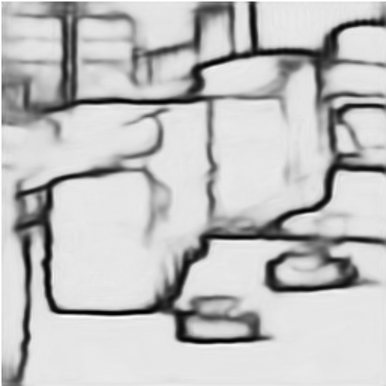}}
    \\[2pt]

    \fbox{\includegraphics[width=0.126\linewidth]{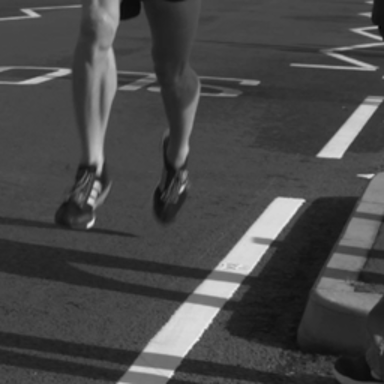}}\hfill
    \fbox{\includegraphics[width=0.126\linewidth]{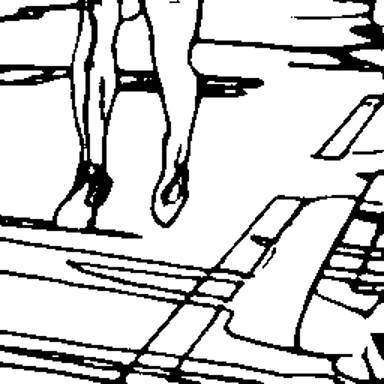}}\hfill
    \fbox{\includegraphics[width=0.126\linewidth]{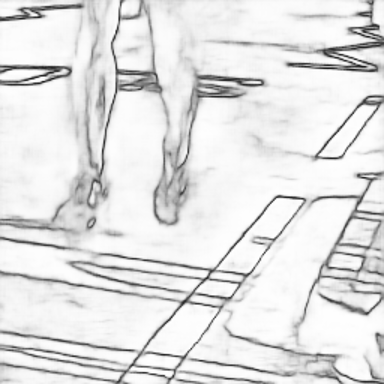}}\hfill
    \fbox{\includegraphics[width=0.126\linewidth]{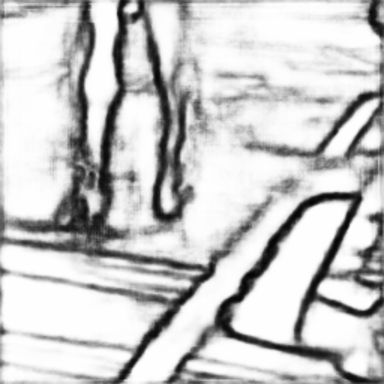}}\hfill
    \fbox{\includegraphics[width=0.126\linewidth]{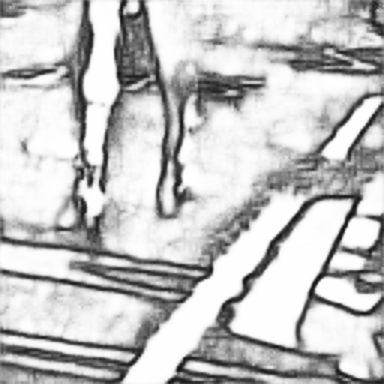}}\hfill
    \fbox{\includegraphics[width=0.126\linewidth]{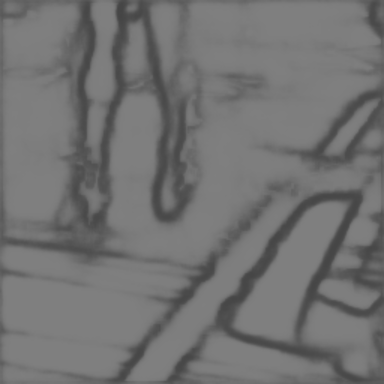}}\hfill
    \fbox{\includegraphics[width=0.126\linewidth]{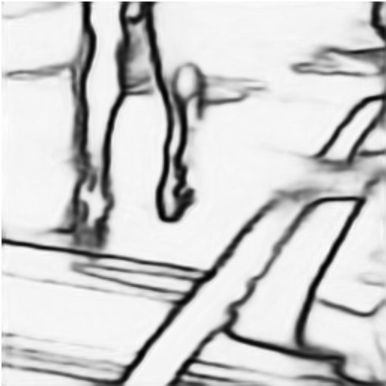}}

    \caption{Qualitative comparison of edge detection results on several benchmark datasets.}
    \label{fig:edge_detection_visual_comparison}
\end{figure*}

\begin{figure*}[t!]
    \centering
    \newlength{\depthimagewidth}
    \setlength{\depthimagewidth}{0.245\linewidth}
    \newlength{\depthimageheight}
    \setlength{\depthimageheight}{0.303\depthimagewidth}

    \begin{minipage}[c]{\depthimagewidth} \centering \footnotesize Clean Frame \end{minipage}\hfill
    \begin{minipage}[c]{\depthimagewidth} \centering \footnotesize GT \end{minipage}\hfill
    \begin{minipage}[c]{\depthimagewidth} \centering \footnotesize \textbf{CompDAE} \end{minipage}\hfill
    \begin{minipage}[c]{\depthimagewidth} \centering \footnotesize DIC+Unidepth \end{minipage}
    \\[2pt]

    \includegraphics[width=\depthimagewidth, height=\depthimageheight]{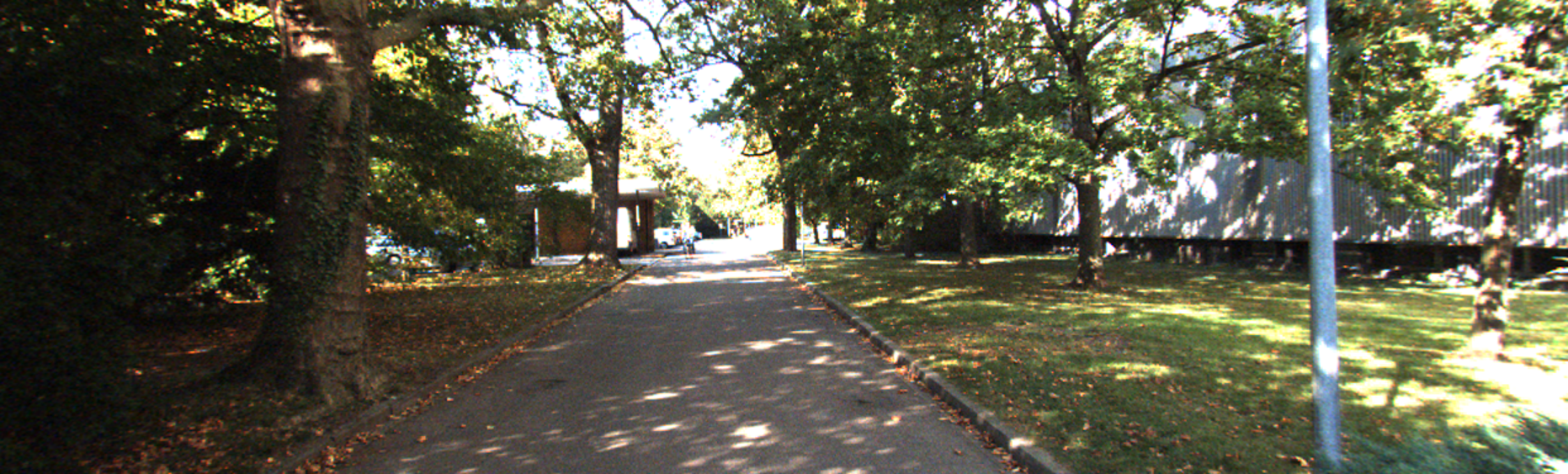}\hfill
    \includegraphics[width=\depthimagewidth, height=\depthimageheight]{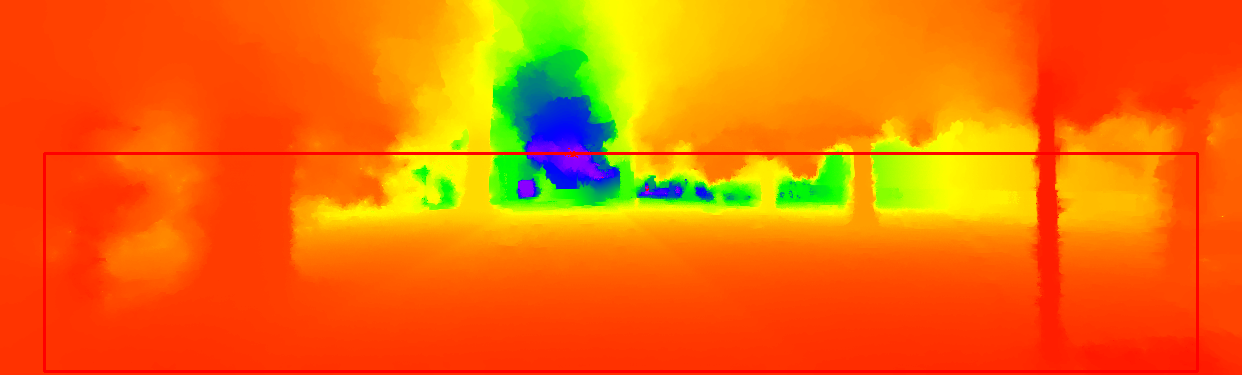}\hfill
    \includegraphics[width=\depthimagewidth, height=\depthimageheight]{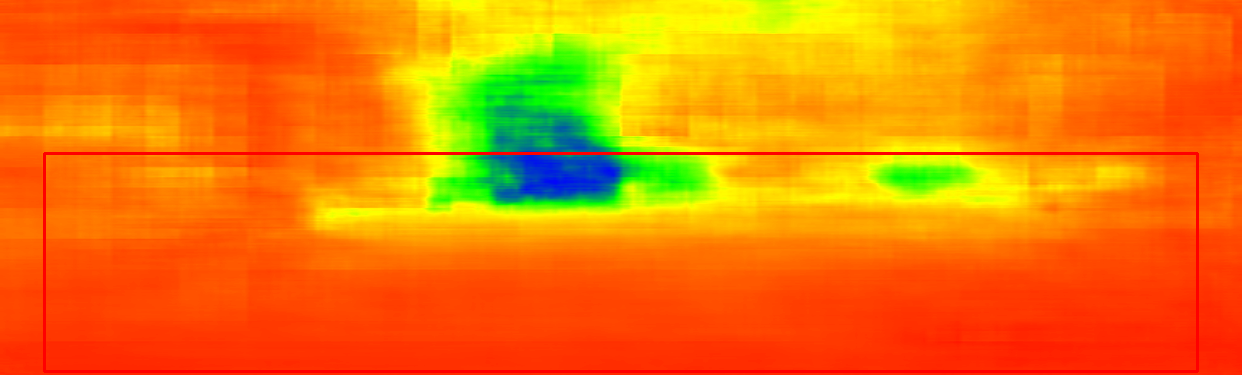}\hfill
    \includegraphics[width=\depthimagewidth, height=\depthimageheight]{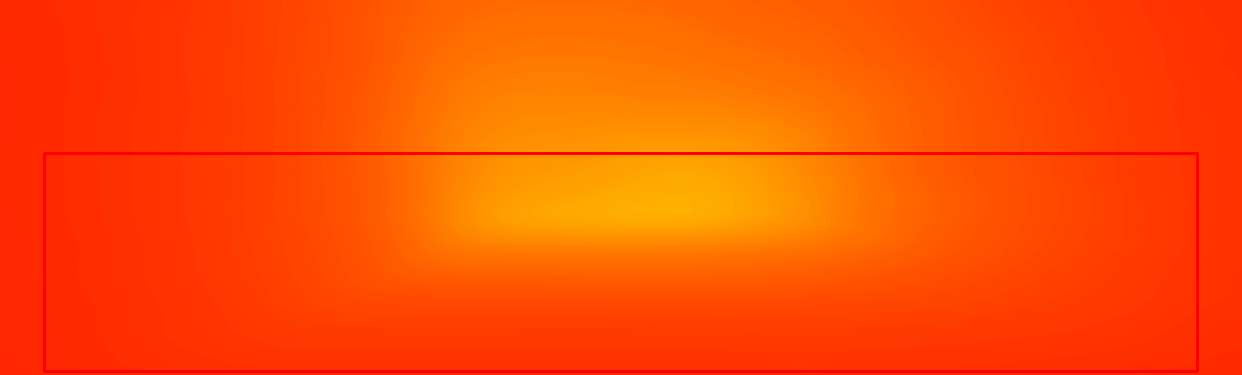}
    \\[2pt]

    \includegraphics[width=\depthimagewidth, height=\depthimageheight]{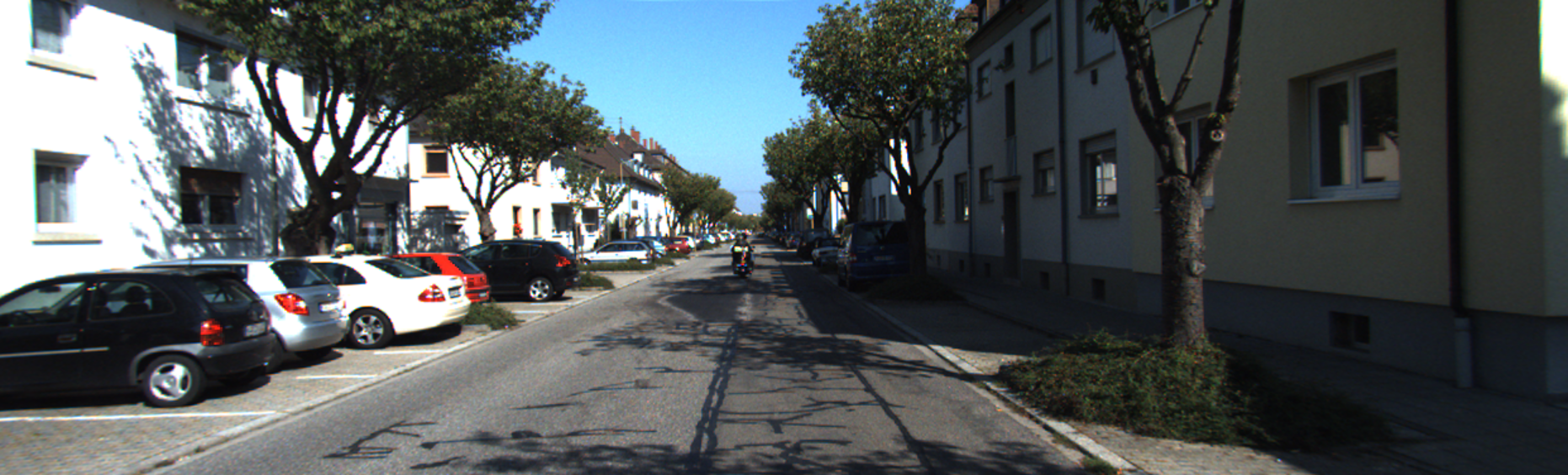}\hfill
    \includegraphics[width=\depthimagewidth, height=\depthimageheight]{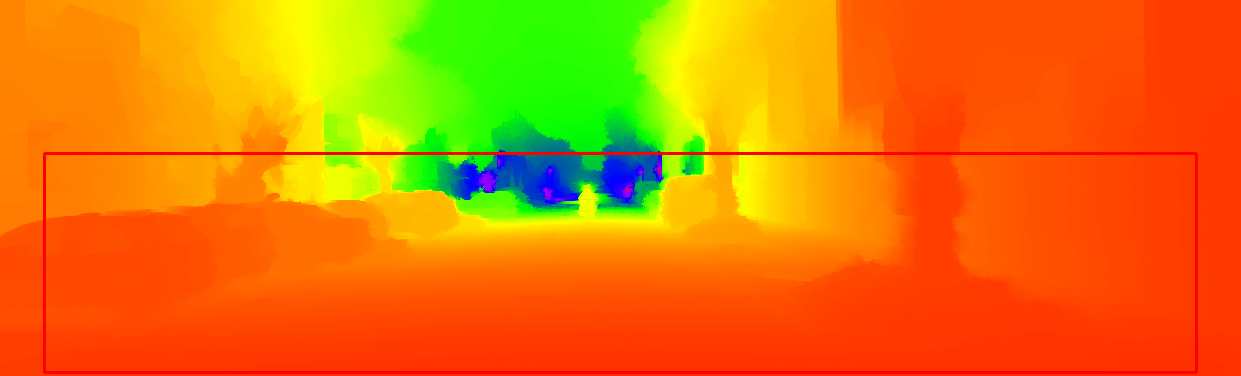}\hfill
    \includegraphics[width=\depthimagewidth, height=\depthimageheight]{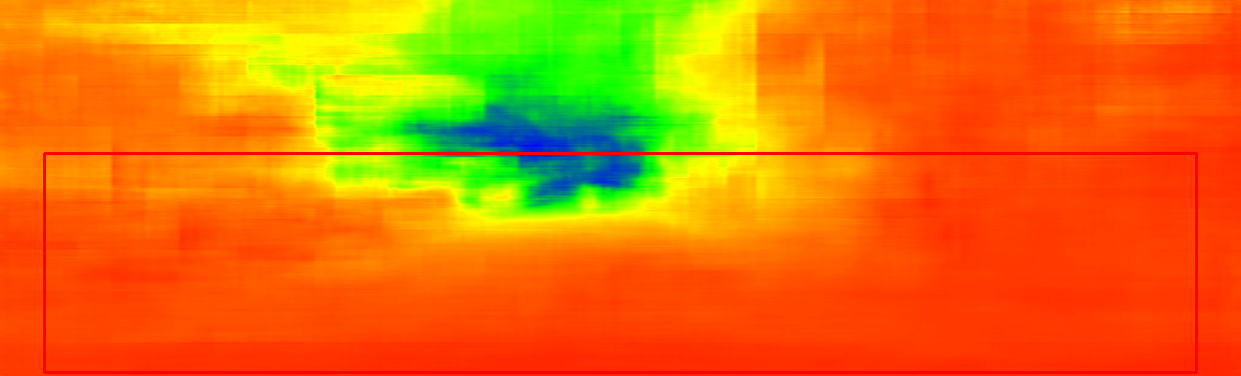}\hfill
    \includegraphics[width=\depthimagewidth, height=\depthimageheight]{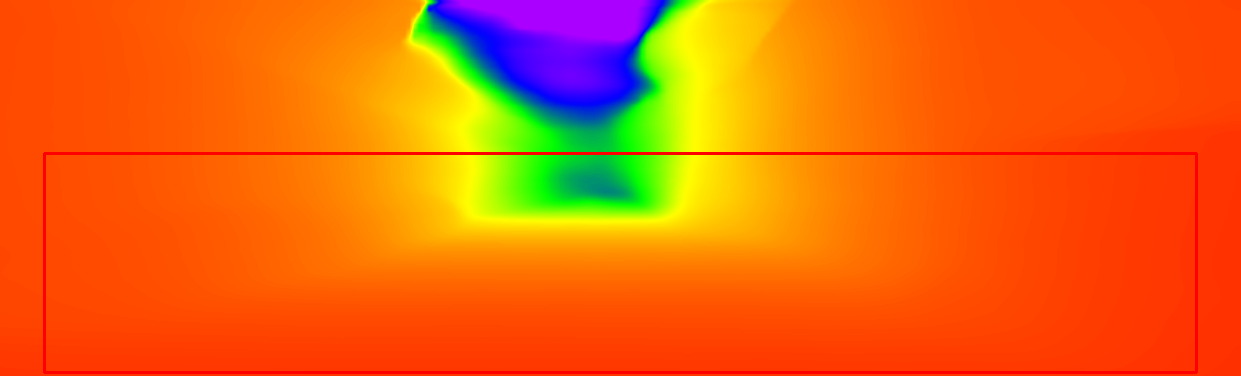}

    \caption{Qualitative comparison of depth estimation results on KITTI test dataset.}
    \label{fig:depth_estimation_visual_comparison}
\end{figure*}

\subsubsection{End-to-End Paradigm vs. Two-Stage Pipeline}

A common SCI vision approach uses a two-stage pipeline: reconstruct the video, then apply a vision model. We compare our end-to-end CompDAE to this baseline, using DIC~\cite{Liu:24} for low-light reconstruction followed by standard task-specific models.
\renewcommand{\arraystretch}{0.8}
\begin{table*}[t!]
    \centering
    \scriptsize
    \begin{subtable}{0.49\textwidth}
        \centering
        \scriptsize
        \begin{tabular}{@{}lcccccc@{}}
            \toprule
            \multirow{2}{*}{Method} & \multicolumn{2}{c}{APC = 5} & \multicolumn{2}{c}{APC = 10} & \multicolumn{2}{c}{APC = 20} \\
            \cmidrule(lr){2-3} \cmidrule(lr){4-5} \cmidrule(lr){6-7}
                    & ODS & OIS & ODS & OIS & ODS & OIS \\ \midrule
            DIC+BDCN   & 0.489 & 0.508 & 0.501 & 0.521 & 0.509 & 0.528 \\
            DIC+DexiNed & 0.460 & 0.473 & 0.507 & 0.517 & 0.547 & 0.554 \\
            DIC+RCF    & 0.471 & 0.497 & 0.492 & 0.521 & 0.498 & 0.528 \\
            DIC+DDN    & 0.433 & 0.484 & 0.459 & 0.496 & 0.487 & 0.509 \\ \midrule
            CompDAE    & \textbf{0.621} & \textbf{0.639} & \textbf{0.650} & \textbf{0.669} & \textbf{0.670} & \textbf{0.685} \\ \bottomrule
        \end{tabular}
        \caption{Edge Detection Performance Comparison.}
        \label{tab:edge_detection_comparison}
    \end{subtable}
    \hfill
    \begin{subtable}{0.49\textwidth}
        \centering
        \scriptsize
        \begin{tabular}{@{}lcccccc@{}}
            \toprule
            Method & AbsRel & RMSE & $\log${10} & $\delta_1$ & $\delta_2$ & $\delta_3$ \\
            \midrule
            DIC+Unidepth & 0.183& 6.514& 0.453& 0.741& 0.842& 0.875 \\
            DIC+MiDaS & 0.415& 7.961& 1.699& 0.337& 0.649& 0.827 \\
            DIC+Metric3D & 0.354& 8.693& 0.359& 0.560& 0.803& 0.892\\
            DIC+LightedDepth & 0.867& 15.742& 0.818& 0.513& 0.602& 0.673\\
            \midrule
            CompDAE & \textbf{0.181}& \textbf{5.177}& \textbf{0.262}& \textbf{0.770}& \textbf{0.893}& \textbf{0.951}\\
            \bottomrule
        \end{tabular}
        \caption{Depth Estimation Performance Comparison.}
        \label{tab:depth_estimation_comparison}
    \end{subtable}
    
    \caption{Quantitative comparison for downstream tasks.}
    \label{tab:downstream_tasks_combined}
\end{table*}

Tables~\ref{tab:edge_detection_comparison} and \ref{tab:depth_estimation_comparison} show that CompDAE consistently outperforms two-stage baselines across edge detection and depth estimation. For semantic segmentation, it achieves 30.43\% mIoU and 82.49\% mPA, far surpassing the DIC+SegFormer baseline (6.60\% mIoU, 45.37\% mPA). Visual comparisons further confirm these gains: CompDAE produces cleaner edges and more accurate depth maps than DIC-based pipelines. These results highlight the effectiveness of our end-to-end design. By extracting task-specific features directly from compressive measurements, CompDAE avoids reconstruction artifacts, improving both accuracy and efficiency while preserving privacy.

\subsubsection{Versatility and Robustness of CompDAE}

Beyond outperforming two-stage pipelines, CompDAE proves to be a versatile and robust general-purpose framework.
CompDAE pairs a shared measurement encoder with lightweight, task-specific decoders to handle edge detection, depth estimation, and semantic segmentation, demonstrating the generality of its learned representations from raw compressive data. Furthermore, as shown in Table~\ref{tab:edge_detection_comparison}, the unified model maintains strong performance across varying low-light conditions (APC = 5, 10, 20), thanks to our unified training strategy that adapts to diverse noise levels.
In summary, CompDAE offers an efficient, end-to-end solution for direct-from-sensor vision across tasks and lighting conditions.

\subsection{Results on Real-World Photon-Limited Video Data}
\begin{figure}[h!]
    \centering
    \begin{subfigure}{0.31\linewidth}
        \centering
        \includegraphics[width=\linewidth]{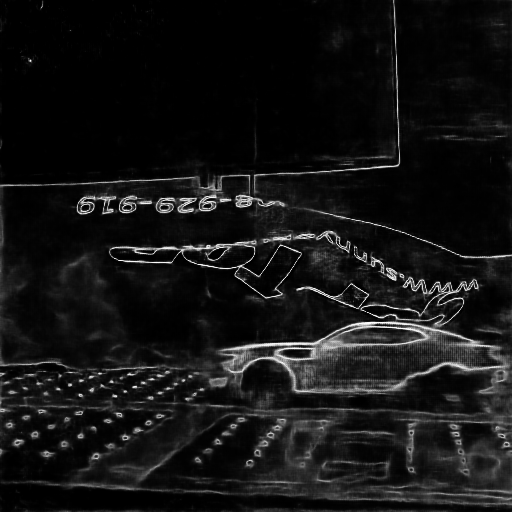}
        \caption{0.1 lux}
        \label{fig:spad_01}
    \end{subfigure}
    \hfill
    \begin{subfigure}{0.31\linewidth}
        \centering
        \includegraphics[width=\linewidth]{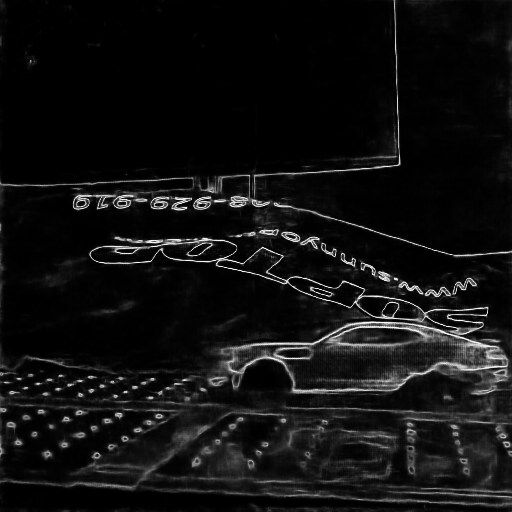}
        \caption{0.2 lux}
        \label{fig:spad_02}
    \end{subfigure}
    \hfill
    \begin{subfigure}{0.31\linewidth}
        \centering
        \includegraphics[width=\linewidth]{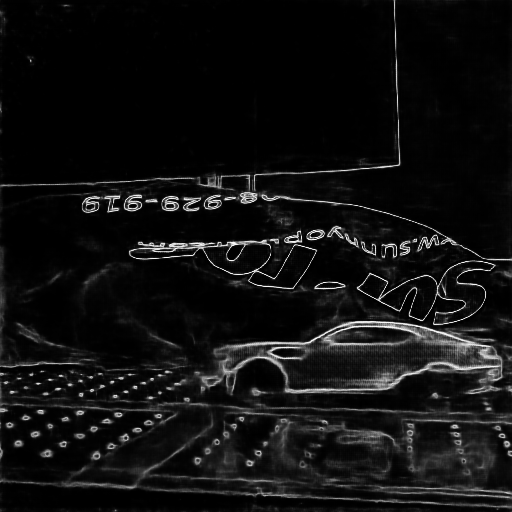}
        \caption{0.4 lux}
        \label{fig:spad_03}
    \end{subfigure}
    \caption{Results of edge detection on the real-world SPAD dataset under different extreme photon-starved conditions.}
    \label{fig:spad_results}
\end{figure}
To assess real-world performance under extreme low-light, we tested CompDAE on video captured by the SPAD512S camera, which provides photon-level imaging at 512×512 resolution and 200 fps. SPAD data is inherently noisy, posing challenges for conventional methods. As shown in Figure~\ref{fig:spad_results}, CompDAE delivers robust results across varied low-light scenes, demonstrating its practical effectiveness and ability to generalize from simulation to real-world photon-limited data.

\section{Discussions}
\label{sec:discussion}

This paper addresses key limitations of traditional SCI systems, which, though high-speed and low-power, struggle under low-light, low-SNR conditions and face hardware challenges due to large frame-sized masks. To address this, we propose a practical design using fixed 8×8 masks, avoiding external memory access and enabling efficient hardware implementation. We introduce CompDAE, a hybrid Transformer-CNN system built on STFormer, which directly processes raw compressive measurements to perform vision tasks like edge detection and depth estimation without full reconstruction. By explicitly modeling noise and leveraging compressed data, CompDAE delivers strong performance in challenging low-light scenarios where conventional methods fail.
Across tasks and datasets, CompDAE matches or exceeds traditional approaches—especially under extreme photon limitations—while significantly reducing model size and complexity. Its modular design supports a universal encoder for sensing and task-specific decoders for diverse vision applications, offering a flexible and scalable framework for future SCI-based vision systems.

Our proposed CompDAE effectively reconstructs multi-frame sequences directly from compressive measurements under low-SNR, photon-limited conditions. By operating in the measurement domain, it avoids full RGB reconstruction, significantly reducing bandwidth and computation—ideal for real-time applications like surveillance, autonomous driving, and mobile devices.
Combining CNNs with spatiotemporal Transformers, CompDAE achieves robust denoising and representation in extreme low-light scenarios. Its privacy-preserving nature—by never producing high-fidelity RGB frames—makes it especially suited for sensitive applications where visual confidentiality is critical.

Despite these strengths, CompDAE uses fixed-sized binary masks, which, while hardware-friendly, limit adaptability across tasks and scenes. Future work may explore adaptive, learnable masks co-optimized with the networks, along with real-time dynamic sensing hardware, enabling more intelligent and flexible compressive vision systems.

\bibliography{aaai2026}

\end{document}